\documentclass[twoside, a4paper, 12pt, bibliography=totoc, titlepage=false]{scrartcl}
\usepackage[shortcuts]{extdash}
\usepackage{graphicx}
\usepackage{color}
\usepackage{pifont}
\usepackage[a4paper]{anysize}
\usepackage[round,colon]{natbib}
\usepackage{times}
\usepackage[hyphens]{url}
\usepackage[german,english]{babel}
\usepackage{moreverb}
\usepackage{amsmath}
\usepackage{amsfonts}
\usepackage{amssymb}
\usepackage{graphicx}
\usepackage{environ}
\usepackage{bm}
\usepackage[utf8]{inputenc}
\usepackage[T1]{fontenc}
\usepackage{nameref}
\marginsize{3.0cm}{3.0cm}{2.0cm}{2.0cm}

\newcommand{\trarxiv}[2]{#2}

\begin{document}
\def\listinglabel#1{\llap{\tiny\ttfamily\the#1}\hskip\listingoffset\relax}

\trarxiv{%
}{%
}

\trarxiv{%
  \titlehead{\includegraphics[width=\textwidth]{logomittext-en_flat_mod.eps}}
}{%
}
\subject{\hspace*{1cm}}
\title{%
  Derivation of Coupled PCA and SVD Learning Rules %
  from a Newton Zero-Finding Framework%
}
\trarxiv{%
  \author{\textsf{Ralf Möller}}
}{%
  \author{%
    Ralf Möller\\
    Computer Engineering Group, Faculty of Technology\\
    Bielefeld University, Bielefeld, Germany\\
    \url{www.ti.uni-bielefeld.de}
  }
}
\trarxiv{%
  \date{\normalsize\textsf{2017\\[5mm]version of \today}}
}{%
  \date{\hspace*{1cm}}
}
\maketitle

\trarxiv{%
  
%
%
%
%
%
%

\newcommand{\summe}[2]{\sum\limits_{#1}^{#2}}
\newcommand{\produkt}[2]{\prod\limits_{#1}^{#2}}
\newcommand{\sumin}{\summe{i=1}{n}}
\newcommand{\sumjn}{\summe{j=1}{n}}
\newcommand{\sumkn}{\summe{k=1}{n}}
\newcommand{\sumim}{\summe{i=1}{m}}
\newcommand{\sumjm}{\summe{j=1}{m}}
\newcommand{\prodin}{\produkt{i=1}{n}}
\newcommand{\prodim}{\produkt{i=1}{m}}
\newcommand{\prodjm}{\produkt{j=1}{m}}
\newcommand{\matA}{\mathbf{A}}
\newcommand{\matB}{\mathbf{B}}
\newcommand{\matC}{\mathbf{C}}
\newcommand{\matD}{\mathbf{D}}
\newcommand{\matE}{\mathbf{E}}
\newcommand{\matF}{\mathbf{F}}
\newcommand{\matG}{\mathbf{G}}
\newcommand{\matH}{\mathbf{H}}
\newcommand{\matI}{\mathbf{I}}
\newcommand{\matJ}{\mathbf{J}}
\newcommand{\matK}{\mathbf{K}}
\newcommand{\matL}{\mathbf{L}}
\newcommand{\matM}{\mathbf{M}}
\newcommand{\matN}{\mathbf{N}}
\newcommand{\matO}{\mathbf{O}}
\newcommand{\matP}{\mathbf{P}}
\newcommand{\matPhi}{\boldsymbol{\Phi}}
\newcommand{\matQ}{\mathbf{Q}}
\newcommand{\matR}{\mathbf{R}}
\newcommand{\matS}{\mathbf{S}}
\newcommand{\matSnull}{\bar\mathbf{S}}
\newcommand{\matT}{\mathbf{T}}
\newcommand{\matU}{\mathbf{U}}
\newcommand{\matUnull}{\bar\mathbf{U}}
\newcommand{\matV}{\mathbf{V}}
\newcommand{\matVnull}{\bar\mathbf{V}}
\newcommand{\matW}{\mathbf{W}}
\newcommand{\matX}{\mathbf{X}}
\newcommand{\matZ}{\mathbf{Z}}
\newcommand{\matWnull}{\bar\mathbf{W}}
\newcommand{\matWdot}{{\dot{\mathbf{W}}}}
\newcommand{\matWe}{{\mathbf{W}^e}}
\newcommand{\matBeta}{\mathbf{B}}
\newcommand{\matDelta}{{\mathbf\Delta}}
\newcommand{\matGamma}{{\mathbf\Gamma}}
\newcommand{\matLambda}{{\mathbf\Lambda}}
\newcommand{\matOmega}{{\mathbf\Omega}}
\newcommand{\matLambdanull}{\bar{\mathbf{\Lambda}}}
\newcommand{\matNull}{\mathbf{0}}
\newcommand{\veca}{\mathbf{a}}
\newcommand{\vecb}{\mathbf{b}}
\newcommand{\vecc}{\mathbf{c}}
\newcommand{\vecd}{\mathbf{d}}
\newcommand{\vece}{\mathbf{e}}
\newcommand{\veceta}{\boldsymbol{\eta}}
\newcommand{\vecf}{\mathbf{f}}
\newcommand{\vecg}{\mathbf{g}}
\newcommand{\vech}{\mathbf{h}}
\newcommand{\veci}{\mathbf{i}}
\newcommand{\vecj}{\mathbf{j}}
\newcommand{\veck}{\mathbf{k}}
\newcommand{\vecl}{\mathbf{l}}
\newcommand{\vecm}{\mathbf{m}}
\newcommand{\vecmu}{\boldsymbol{\mu}}
\newcommand{\vecn}{\mathbf{n}}
\newcommand{\vecnu}{\boldsymbol{\nu}}
\newcommand{\veco}{\mathbf{o}}
\newcommand{\vecp}{\mathbf{p}}
\newcommand{\vecphi}{\boldsymbol{\varphi}}
\newcommand{\vecq}{\mathbf{q}}
\newcommand{\vecr}{\mathbf{r}}
\newcommand{\vecs}{\mathbf{s}}
\newcommand{\vect}{\mathbf{t}}
\newcommand{\vecu}{\mathbf{u}}
\newcommand{\vecunull}{\bar\mathbf{u}}
\newcommand{\vecv}{\mathbf{v}}
\newcommand{\vecvnull}{\bar\mathbf{v}}
\newcommand{\vecx}{\mathbf{x}}
\newcommand{\vecxdot}{\dot{\mathbf{x}}}
\newcommand{\vecxxi}{\boldsymbol{\xi}}
\newcommand{\vecy}{\mathbf{y}}
\newcommand{\vecz}{\mathbf{z}}
\newcommand{\veczeta}{\boldsymbol{\zeta}}
\newcommand{\vecw}{\mathbf{w}}
\newcommand{\vecwe}{{\mathbf{w}^e}}
\newcommand{\vecwei}{{\mathbf{w}^e_i}}
\newcommand{\vecwek}{{\mathbf{w}^e_k}}
\newcommand{\vecwdot}{\dot{\mathbf{w}}}
\newcommand{\vecwnull}{\bar\mathbf{w}}
\newcommand{\vecdwdt}{\frac{d\vecw}{dt}}
\newcommand{\vecnull}{\mathbf{0}}
\newcommand{\vecone}{\mathbf{1}}
\newcommand{\vecvdot}{\dot{\mathbf{v}}}
\newcommand{\vecomega}{\boldsymbol{\omega}}
\newcommand{\norm}[1]{\|#1\|}
\newcommand{\normF}[1]{\|#1\|_F}
\newcommand{\normm}{\norm{\vecm}}
\newcommand{\normw}{\norm{\vecw}}
\newcommand{\normx}{\norm{\vecx}}
\newcommand{\xp}[1]{\langle #1\rangle}
\newcommand{\xpvecm}{\xp{\vecm}}
\newcommand{\normxpm}{\norm{\xpvecm}}
\newcommand{\lambdanull}{\bar{\lambda}}
\newcommand{\Wnull}{\bar{W}}
\newcommand{\half}{\frac{1}{2}}
\newcommand{\third}{\frac{1}{3}}
\newcommand{\order}[1]{{\cal O}(#1)}
\newcommand{\ddt}{\frac{d}{dt}}
\newcommand{\vecstk}[1]{\left(\begin{array}{c}#1\end{array}\right)}
\newcommand{\blkstk}[1]{\left(\begin{array}{c|c}#1\end{array}\right)}
\newcommand{\pmat}[1]{\begin{pmatrix}#1\end{pmatrix}}
%
\def\tr{\mathop{\rm tr}\nolimits}
\def\diag{\mathop{\rm diag}\nolimits}
\def\cov{\mathop{\rm cov}\nolimits}
\def\var{\mathop{\rm var}\nolimits}
\newcommand{\rank}{\operatorname{rank}}
\newcommand{\sgn}{\operatorname{sgn}}
\newcommand{\atantwo}{\operatorname{atan2}}
\newcommand{\asin}{\operatorname{asin}}
\newcommand{\acos}{\operatorname{acos}}
\newcommand{\atan}{\operatorname{atan}}
\newcommand{\adj}{\operatorname{adj}}

\newcommand{\ddf}[2]{\frac{\partial #1}{\partial #2}}
\newcommand{\df}[1]{\ddf{}{#1}}
\newcommand{\ddfs}[3]{\frac{\partial^2 #1}{\partial #2\partial #3}}
\newcommand{\ddfsq}[2]{\frac{\partial^2 #1}{\partial {#2}^2}}
\newcommand{\pihalf}{\frac{\pi}{2}}

\newcommand{\grad}{\nabla}
\newcommand{\gradv}{\boldsymbol{\nabla}}

\newcommand{\eps}{\varepsilon}

}{%
  
}
\newcommand{\mattW}{\tilde{\matW}}
\newcommand{\vectw}{\tilde{\vecw}}
\newcommand{\mattU}{\tilde{\matU}}
\newcommand{\vectu}{\tilde{\vecu}}
\newcommand{\mattV}{\tilde{\matV}}
\newcommand{\vectv}{\tilde{\vecv}}


\begin{abstract}
\noindent\sloppy In coupled learning rules for PCA (principal
component analysis) and SVD (singular value decomposition), the update
of the estimates of eigenvectors or singular vectors is influenced by
the estimates of eigenvalues or singular values, respectively. This
coupled update mitigates the speed-stability problem since the update
equations converge from all directions with approximately the same
speed. A method to derive coupled learning rules from information
criteria by Newton {\em optimization} is known. However, these
information criteria have to be designed, offer no explanatory value,
and can only impose Euclidean constraints on the vector
estimates. Here we describe an alternative approach where coupled PCA
and SVD learning rules can systematically be derived from a Newton
{\em zero-finding} framework. The derivation starts from an objective
function, combines the equations for its extrema with arbitrary
constraints on the vector estimates, and solves the resulting vector
zero-point equation using Newton's zero-finding method. To demonstrate
the framework, we derive PCA and SVD learning rules with constant
Euclidean length or constant sum of the vector estimates.\\[0.5cm]
\trarxiv{%
  Please cite as: Ralf Möller. {\em Derivation of Coupled PCA and SVD
    Learning Rules from a Newton Zero-Finding Framework}. Technical
  Report, Computer Engineering, Faculty of Technology, Bielefeld
  University, 2017, version of \today, \url{www.ti.uni-bielefeld.de}.
}{%
}
\end{abstract}


\thispagestyle{empty}
\newpage
\tableofcontents
\thispagestyle{empty}
\newpage
\setcounter{page}{1}


\sloppypar

\section{Introduction}

Coupled learning rules have been developed to mitigate the
speed-stability problem in online learning rules for principal
component analysis (PCA) or singular value decomposition (SVD)
\cite[for reviews see][]{own_Moeller04a,own_Kaiser10}. Coupled
learning rules are systems of ordinary differential equations (ODEs)
where not only the principal eigenvectors or singular vectors are
estimated (vector estimates), but simultaneously also the principal
eigenvalues or singular values (scalar estimates). The ODEs for vector
and scalar estimates are coupled, and it is the influence of the
scalar estimates on the ODEs of the vector estimates that ensures fast
convergence to the stationary points from all directions.

As we have suggested earlier \cite[]{own_Moeller04a,own_Kaiser10},
coupled learning rules can be derived by applying a Newton descent
\begin{equation}
\vecxdot = -\matH^{-1}(\vecx) \ddf{p(\vecx)}{\vecx}
\end{equation}
to an information criterion $p(\vecx)$. The Hessian matrix
$\matH(\vecx)$ of $p(\vecx)$ has to be analytically inverted in the
vicinity of the desired stationary point (e.g. at the principal
eigenvector / eigenvalue pair). The information criterion only has to
have the desired stationary points, regardless of whether they are
attractors, repellers, or saddle points. The Newton descent at the
desired stationary point turns this stationary point into an attractor
and leads to equal convergence speed from all directions.

We have proposed \cite[]{own_Moeller04a} the following information
criterion for the derivation of coupled learning rules which extract
the principal or minor eigenvector / eigenvalue pair $\vecx^T = (\vecw^T |
\lambda)$ from a covariance matrix $\matC$:
\begin{equation}\label{eq_p_pca1}
p(\vecw,\lambda) = 
\vecw^T \matC \vecw \lambda^{-1} - \vecw^T \vecw + \ln \lambda.
\end{equation}
The same learning rules can also be derived from another criterion
suggested by \cite{nn_Hou06} (original publication not available to
us, cited after \cite{nn_Feng17}):
\begin{equation}\label{eq_p_pca2}
p(\vecw,\lambda) = 
\vecw^T \matC \vecw - \vecw^T \vecw \lambda + \lambda.
\end{equation}
The resulting online learning rule for the vector estimate resembles
Oja's rule \cite[]{nn_Oja82} with an additional factor $\lambda^{-1}$
that influences the effective learning rate; the coupled version
resembles ``ALA'' \cite[]{nn_Chen95}.

For singular value decomposition, coupled learning rules for the
principal singular vectors / singular value triplet $\vecx^T = (\vecu^T |
\vecv^T | \sigma)$ of a cross-covariance matrix $\matA$ can be obtained
from the information criterion
\begin{equation}\label{eq_p_svd1}
p(\vecu, \vecv, \sigma) 
= \vecu^T \matA \vecv \sigma^{-1} - \half\vecu^T \vecu 
- \half \vecv^T \vecv + \ln \sigma
\end{equation}
as suggested by \cite{own_Kaiser10}. Supposedly an alternative similar
to the PCA criterion by \cite{nn_Hou06} also exists for the SVD
case. The online learning rules derived from this criterion for the
vector estimates resemble the ``cross-coupled Hebbian rule'' suggested
by \cite{nn_Diamantaras94} with an additional factor $\sigma^{-1}$.

For the generalized eigenproblem $\matR_y \vecw = \lambda \matR_x
\vecw$, similar information criteria have been proposed by
\cite{nn_Nguyen13}
\begin{equation}\label{eq_p_gpca1}
p(\vecw,\lambda) 
= \vecw^H \matR_y \vecw \lambda^{-1} - \vecw^H \matR_x \vecw + \ln \lambda
\end{equation}
and by \cite{nn_Feng16a}
\begin{equation}\label{eq_p_gpca2}
p(\vecw,\lambda)
= \vecw^H \matR_y \vecw - \vecw^H \matR_x \vecw \lambda + \lambda.
\end{equation}

The approach of deriving learning rules from an information criterion
by a Newton descent (the latter being commonly used in optimization
problems) has obviously proven its value, but is limited in three
ways:
\begin{enumerate}
\item An information criterion has to be designed that has the desired
  stationary points. While the design is simplified by the fact that
  the relevant stationary point doesn't have to be an attractor, there
  is currently no systematic way to obtain such a criterion.
\item The information criterion has no explanatory value. The decisive
  property is just that is has the desired stationary points, but the
  criterion doesn't reveal anything about the problem at hand since
  the desired stationary point is typically not an attractor.
\item The PCA and SVD information criteria listed above
  (\ref{eq_p_pca1},\ref{eq_p_pca2},\ref{eq_p_svd1}) lead to solutions
  where the vector estimates have a Euclidean (L2) norm of
  $1$. Information criteria where the vector estimates fulfill other
  constraints in the stationary point are currently not known.
\end{enumerate}
In this paper we suggest an alternative approach which resolves these
limitations. Instead of deriving learning rules from a Newton {\em
  descent}, we use a Newton {\em zero-finder} to find the zero points
of systems of equations. These equations are easy to derive (e.g. by
optimizing some objective function), they are directly related to the
problem (e.g. they constitute the well-known eigen equations), and
different constraints can be imposed on the vector estimates by adding
the appropriate equations.

\section{Newton Zero-Finding Framework}

Given an equation $\vecf(\vecx) = \vecnull$, the Newton zero-finder
ODE is given by
\begin{equation}\label{eq_newton_zero}
\vecxdot = -\matJ^{-1}(\vecx) \vecf(\vecx)
\end{equation}

where $\matJ(\vecx)$ is the Jacobian matrix of $\vecf(\vecx)$. In a
similar way as in our earlier paper \cite[][appendix
  I]{own_Moeller04a} we can show that the speed of convergence is the
same from all directions: If we insert the first-order Taylor
approximation of $\vecf(\vecx)$ at the zero point $\vecx_0$,
\begin{equation}
\vecf(\vecx) = \vecf(\vecx_0) + \matJ(\vecx_0) \cdot (\vecx - \vecx_0) + \ldots
\end{equation}
into equation (\ref{eq_newton_zero}) and take into account that
$\vecf(\vecx_0) = \vecnull$, we obtain
\begin{equation}\label{eq_xdot_tmp1}
\vecxdot = -\matJ^{-1}(\vecx) 
\left[\matJ(\vecx_0) \cdot (\vecx - \vecx_0) + \ldots \right].
\end{equation}
If we also approximate $\matJ^{-1}(\vecx)$ in a first-order Taylor expansion as
\begin{equation}\label{eq_Jinv_taylor}
\matJ^{-1}(\vecx) = \matJ^{-1}(\vecx_0) + \mathcal{O}(\vecx - \vecx_0) + \ldots
\end{equation}
and omit second-order terms after inserting (\ref{eq_Jinv_taylor})
into (\ref{eq_xdot_tmp1}), we get
\begin{equation}
\vecxdot = -(\vecx - \vecx_0).
\end{equation}
This ODE has an attractor in $\vecx_0$ and converges with the same
speed from all directions.

In the following, we will start the derivation of each ODE system from
some objective function. This preparatory step proved to be necessary
for the SVD system with non-Euclidean constraint on the weight vectors
since the well-known SVD equations (as for example used by
\cite{own_Kaiser10}) only apply for a Euclidean constraint. The
objective function is then turned into a zero-finding problem
formulated over the vector estimates and scalar estimates. The desired
constraints are added and the learning rules are derived from
(\ref{eq_newton_zero}), in a way similar to our earlier approach
\cite[]{own_Moeller04a,own_Kaiser10}. Online forms of the rules can
finally be derived by replacing the covariance / cross-covariance
matrices by rank-1 outer vector products.

We derive equations for PCA with Euclidean weight vector norm
(reproducing the results by \cite{own_Moeller04a}), for PCA with
constant weight vector sum (new), for SVD with Euclidean weight vector
norm (similar to the derivation by \cite{own_Kaiser10}), and for SVD
with constant weight vector sum (new).

All four derivations go through the following steps:
\begin{enumerate}
\item Define an objective function independent of the length of the
  vector estimate.
\item Determine the optimum of the objective function.
\item Introduce scalar estimates.
\item Define the zero-point problem by adding constraints on the
  vector estimates.
\item Compute the Jacobian of the zero-point function.
\item Apply an orthogonal transformation to the Jacobian.
\item Interrelate between the vector estimates in Euclidean norm and
  the given constraint.
\item Approximate the transformed Jacobian for the desired zero point.
\item Invert the approximated transformed Jacobian (e.g. by
  Gauss-Jordan elimination).
\item Apply the inverted orthogonal transformation.
\item Extract the ODEs for vector estimates and scalar estimates.
\item Compute the online ODEs for vector estimates and scalar
  estimates.
\end{enumerate}

\section{PCA}

\subsection{PCA Objective Function}

The objective of PCA is to find a weight vector $\vecw$ which maximizes
the variance of the projection of a vector $\vecx$ (drawn from a
random distribution) onto this weight vector. We define the projection
as
\begin{equation}
\hat{\xi} = \frac{\vecw^T}{\|\vecw\|}\vecx
\end{equation}
and the objective function as variance of the projection:
\begin{equation}
p(\vecw) = \half E\{\hat{\xi}^2\}.
\end{equation}
We see that
\begin{eqnarray}
p(\vecw) 
&=& 
\half E
\left\{\frac{\vecw^T}{\|\vecw\|}\vecx\vecx^T\frac{\vecw}{\|\vecw\|} \right\}\\
&=&
\half \frac{\vecw^T}{\|\vecw\|} E\{\vecx\vecx^T\} \frac{\vecw}{\|\vecw\|}\\
&=&
\label{eq_rayleigh}
\half \frac{\vecw^T \matC \vecw}{\vecw^T \vecw}
\end{eqnarray}
where $\matC = E\{\vecx\vecx^T\}$ is the covariance matrix of
$\vecx$. Equation (\ref{eq_rayleigh}) is the well-known Rayleigh
quotient.

The derivative of the Rayleigh quotient for a symmetric matrix is
given by equation (\ref{eq_derivative_rayleigh_symm}) in appendix
\ref{app_rayleigh}. We obtain
\begin{equation}
\ddf{p(\vecw)}{\vecw}
=
\frac{1}{\vecw^T\vecw}
\left(\matC\vecw - \vecw \frac{\vecw^T\matC\vecw}{\vecw^T\vecw}\right).
\end{equation}
The extreme point of this equation is given by
\begin{eqnarray}
\frac{1}{\vecw^T\vecw}
\left(\matC\vecw - \vecw \frac{\vecw^T\matC\vecw}{\vecw^T\vecw}\right) 
&=& \vecnull\\
\matC\vecw - \vecw \frac{\vecw^T\matC\vecw}{\vecw^T\vecw} &=& \vecnull.
\end{eqnarray}
The next step is crucial for the derivation of coupled learning rules
as it introduces the scalar estimate, in this case the eigenvalue. We
define
\begin{equation}\label{eq_pca_lambda}
\lambda = \frac{\vecw^T\matC\vecw}{\vecw^T\vecw}
\end{equation}
and obtain the well-known PCA equation to which the Newton zero-finder
is applied below:
\begin{equation}\label{eq_pca}
\matC \vecw = \lambda \vecw.
\end{equation}
By inserting (\ref{eq_pca}) into (\ref{eq_pca_lambda}) we can verify
that this replacement is consistent. It is currently unclear whether
the replacement of a scalar sub-expression by a variable which becomes
part of the solution vector is generally applicable or can only be
used for cases like PCA or SVD equations.

\subsection{PCA with Euclidean Weight Vector Norm}\label{sec_pca_l2}

We can now combine the PCA equation (\ref{eq_pca}) with an Euclidean
(L2) constraint on the weight vector to define the following equation
over the vector $\vecz^T = (\vecw^T | \lambda)$ (we use $\vecz$ here
since $\vecx$ is reserved for input vectors):
\begin{equation}\label{eq_f_pca_l2}
\vecf(\vecz) = \vecf(\vecw,\lambda) =
\pmat{\matC \vecw - \lambda \vecw\\ \half(\vecw^T\vecw - 1)}
\end{equation}
The zero points of this equation are all unit-length eigenvectors and
eigenvalues of $\matC$. The Jacobian of this system is
\begin{equation}
\matJ(\vecw,\lambda) =
\ddf{\vecf(\vecz)}{\vecz} = 
\pmat{
\matC - \lambda \matI & -\vecw\\
\vecw^T & 0
}
\end{equation}
The Jacobian needs to be inverted in the vicinity of the desired root,
which for PCA is the {\em principal} eigenvector / eigenvalue
pair. Inversion in the vicinity of the desired root requires an
orthogonal transformation of the Jacobian into
\begin{equation}\label{eq_trans_fwd}
\matJ^* = \matT^T \matJ \matT. 
\end{equation}
For the PCA case, we use
\begin{equation}\label{eq_T_pca}
\matT = \pmat{
\mattW & \vecnull\\
\vecnull^T & 1
},
\end{equation}
where $\mattW$ contains all unit-length eigenvectors
$\vectw_i$ of $\matC$ in its columns. The matrix $\mattW$ is
orthogonal for disjunct non-zero eigenvalues, i.e. $\mattW^T\mattW =
\mattW\mattW^T = \matI$, and thus also $\matT$ is orthogonal,
i.e. $\matT^T \matT = \matT\matT^T = \matI$. This Jacobian is inverted
and transformed back by
\begin{equation}\label{eq_trans_inv}
\matJ^{-1} = \matT {\matJ^*}^{-1} \matT^T.
\end{equation}
The transformed Jacobian $\matJ^*$ can be approximated in the vicinity
of the principal eigenvector $\vectw_1$ for which the corresponding
eigenvalue $\lambda_1$ is much larger than all other eigenvalues
($\lambda_1 \gg \lambda_i \;\forall i \neq 1$). This step selects the
zero point which we want to approach. We approximate $\vecw \approx
\vectw_1$ and $\lambda \approx \lambda_1$. From the eigen equations
$\matC \mattW = \mattW \matLambda$ and from $\mattW^T\vecw \approx
\vece = (1, 0, \ldots, 0)^T$ we obtain
\begin{equation}
\matJ^* = \pmat{
\matLambda - \lambda \matI & -\vece\\
\vece^T & 0
}.
\end{equation}
In the vicinity of the principal eigenvector / eigenvalue pair we can
approximate
\begin{equation}
\matLambda - \lambda \matI = \pmat{
\lambda_1 - \lambda &                     &        &                     \\
                    & \lambda_2 - \lambda &        &                     \\
                    &                     & \ddots &                     \\    
                    &                     &        & \lambda_n - \lambda
}
\approx \pmat {
0 &          &        &         \\
  & -\lambda &        &         \\
  &          & \ddots &         \\
  &          &        & -\lambda
}
=
\lambda (\vece\vece^T - \matI)
\end{equation}
where $n$ is the dimension of the input vectors. This gives
\begin{equation}
\matJ^* \approx \pmat
{
\lambda (\vece\vece^T - \matI) & -\vece\\
\vece^T & 0
}.
\end{equation}
Inversion of $\matJ^*$ is most easily done by writing the matrix out
as single elements and using Gauss-Jordan elimination to transform
$(\matJ^*|\matI)$ via exchange of rows, scaling of rows, or addition
of scaled rows into $(\matI|{\matJ^*}^{-1})$. For this case we obtain
\begin{equation}
{\matJ^*}^{-1} \approx \pmat
{
\lambda^{-1} (\vece\vece^T-\matI) & \vece\\
-\vece^T & 0
}.
\end{equation}
The test of whether $\matJ^* {\matJ^*}^{-1} = \matI$ holds can easily
be done by block-wise matrix multiplication in vector notation (rather
than by multiplication in single-element notation).

Now the matrix is transformed back using equation
(\ref{eq_trans_inv}). We approximate $\mattW \vece = \vectw_1 \approx
\vecw$ and obtain
\begin{equation}\label{eq_Jinv_pca_l2}
\matJ^{-1}(\vecw,\lambda) \approx \pmat{
\lambda^{-1}(\vecw\vecw^T-\matI) & \vecw\\
-\vecw^T & 0
}
\end{equation}
Finally we compute the ODE system from equation (\ref{eq_newton_zero})
\begin{equation}\label{eq_ode_pca}
\pmat{\vecwdot\\\dot{\lambda}} = -\matJ^{-1}(\vecw,\lambda)\vecf(\vecw,\lambda)
\end{equation}
into which we insert (\ref{eq_Jinv_pca_l2}) and (\ref{eq_f_pca_l2})
\begin{equation}
\pmat{\vecwdot\\\dot{\lambda}} = 
-\pmat{
\lambda^{-1}(\vecw\vecw^T-\matI) & \vecw\\
-\vecw^T & 0
}
\pmat{\matC \vecw - \lambda \vecw\\ \half(\vecw^T\vecw - 1)}.
\end{equation}
This leads to the learning rule ODEs
\begin{eqnarray}
\vecwdot \label{eq_wruleavg_pca_l2}
&=& \lambda^{-1} (\matC\vecw - (\vecw^T\matC\vecw)\vecw) 
+ \half(\vecw^T\vecw-1)\vecw\\
\dot{\lambda} \label{eq_lruleavg_pca_l2}
&=&
\vecw^T\matC\vecw - \lambda\vecw^T\vecw
\end{eqnarray}
which coincide with the ``nPCA'' rules derived by
\cite{own_Moeller04a}.

Online rules can be derived by replacing $\matC$ with $\vecx\vecx^T$
where $\vecx$ is a data vector; the computation of the expectation
$E\{\vecx\vecx^T\}$ is transferred to the averaging properties of the
learning rule. If we introduce the neuron's activity as $\xi =
\vecw^T\vecx$, we get
\begin{eqnarray}
\vecwdot
&=&
\lambda^{-1}\xi(\vecx - \xi\vecw) + \half(\vecw^T\vecw-1)\vecw\\
\dot{\lambda}
&=&
\xi^2 - \vecw^T\vecw\lambda.
\end{eqnarray}
We recognize the resemblance to Oja's L2 rule $\vecwdot = \xi(\vecx -
\xi\vecw)$ which was derived from approximating a normalization to
unit length for small learning rates \cite[]{nn_Oja82}. The factor
$\lambda^{-1}$ ensures fast convergence. If we approximate
$\vecw^T\vecw\approx 1$ in the vicinity of the solution, we obtain the
``ALA'' system suggested by \cite{nn_Chen95}.

\subsection{PCA with Constant Weight Vector Sum}

If we demand that the sum of the elements of the weight vector is
constant (unity), we start from the zero-point function
\begin{equation}
\vecf(\vecz) = \vecf(\vecw,\lambda) =
\pmat{\matC \vecw - \lambda \vecw\\ \vecone^T\vecw-1},
\end{equation}
where $\vecone^T = (1,1,\ldots,1)$. The Jacobian of this function is
\begin{equation}
\matJ(\vecw,\lambda) = \ddf{\vecf(\vecz)}{\vecz} = \pmat{
\matC - \lambda\matI & -\vecw\\
\vecone^T & 0
}
\end{equation}
The orthogonal transformation (\ref{eq_trans_fwd}) is done by the same
transformation matrix (\ref{eq_T_pca}). However, we now have to
establish a relationship between the L2 unit-length vectors in
$\mattW$ and the weight vector $\vecw$ which in the zero point
is constrained to constant sum. We obtain the relationships
\begin{equation}\label{eq_w_wt}
\vecw_i = \frac{\vectw_i}{\vecone^T\vectw_i},\quad
\vectw_i = \frac{\vecw_i}{\|\vecw_i\|},\quad
\mbox{thus} \quad \|\vecw_i\|\cdot(\vecone^T\vectw_i) = 1
\end{equation}
which can be verified by showing that $\vecone^T\vecw_i=1$ and
$\vectw_i^T\vectw_i=1$, respectively.\footnote{\label{fn_w_wt}Note
  that none of the vectors should be parallel to the constant-sum
  plane, since then $\vecone^T\vectw_i = 0$. This should be guaranteed
  if $\vecw_1$ is not a multiple of $\vecone$.} We approximate $\vecw
\approx \vecw_1$ for $\lambda \approx \lambda_1 \gg \lambda_j\;\forall
i \neq 1$. With
\begin{equation}
\mattW^T \vecw = \mattW^T \frac{\vecw}{\|\vecw\|}\|\vecw\| =
\mattW^T \vectw \|\vecw\| = \vece \|\vecw\|
\end{equation}
we get the transformed Jacobian and approximate in the vicinity of the
desired zero point:
\begin{equation}
\matJ^* = \pmat{
\matLambda - \lambda \matI & -\vece \|\vecw\|\\
\vecone^T\mattW & 0
} \approx \pmat{
\lambda(\vece\vece^T-\matI) & -\vece \|\vecw\|\\
\vecone^T\mattW & 0
}.
\end{equation}
For the single-element Gauss-Jordan elimination we introduce the vector
$\vecs^T = \vecone^T\mattW = (s_1,\ldots,s_n)$. The inversion gives
\begin{equation}
{\matJ^*}^{-1} \approx \pmat{
\lambda^{-1}[(\vece\vece^T-\matI)+s_1^{-1}\vece (0,s_2,\ldots,s_n)] 
  & s_1^{-1}\vece\\
-\vece^T{\|\vecw\|}^{-1} & 0
}.
\end{equation}
The inverse orthogonal transformation via (\ref{eq_trans_inv}) requires the
following computation for the second term of the upper-left element:
\begin{eqnarray}
&&(0,s_2,\ldots,s_n)\mattW^T\\ 
&=& (s1,\ldots,s_n)\mattW^T - s_1 \vece^T \mattW^T\\
&=& \vecone^T\mattW\mattW^T - s_1 \vectw^T\\
&=& \vecone^T - \vecone^T\vectw \vectw^T\\
&=& \vecone^T(\matI - \vectw\vectw^T)
\end{eqnarray}
Moreover, we have $\mattW s_1^{-1}\vece = s_1^{-1}\vectw = \vecw$,
such that the upper-left element becomes
\begin{eqnarray}
&&
\lambda^{-1}[(\vectw\vectw^T-\matI)+\vecw\vecone^T(\matI-\vectw\vectw^T)]\\
&=& 
\lambda^{-1}[(\vectw\vectw^T-\matI)
+\vecw\vecone^T
-\vecw(\vecone^T\vectw)\vectw^T)\\
&=&
\lambda^{-1}[(\vectw\vectw^T-\matI)
+\vecw\vecone^T
-\vectw\vectw^T)\\
&=&
\lambda^{-1}(\vecw\vecone^T-\matI).
\end{eqnarray}
For the lower-left element we see that 
\begin{equation}
\vece^T\|\vecw\|^{-1}\mattW^T =
{\|\vecw\|}^{-1}\vectw^T={\|\vecw\|}^{-1}{\|\vecw\|}^{-1}\vecw^T =
{(\vecw^T\vecw)}^{-1}\vecw^T,
\end{equation}
and for the upper-right element we also have $s_1^{-1}\mattW\vece =
s_1^{-1}\vectw = \vecw$, so the inverted Jacobian becomes
\begin{equation}
\matJ^{-1}(\vecw,\lambda) \approx \pmat{
\lambda^{-1}(\vecw\vecone^T-\matI) & \vecw\\
-\frac{\vecw}{\vecw^T\vecw} & 0
}.
\end{equation}
From (\ref{eq_ode_pca}) we get
\begin{eqnarray}
\vecwdot 
&=& 
-[\lambda^{-1}(\vecw\vecone^T-\matI)(\matC\vecw-\lambda\vecw)
+\vecw(\vecone^T\vecw-1)]\\
&=& \label{eq_wruleavg_pca_sum}
\lambda^{-1}(\matC\vecw-(\vecone^T\matC\vecw)\vecw)\\
\dot{\lambda} \label{eq_lruleavg_pca_sum}
&=& \frac{\vecw^T\matC\vecw}{\vecw^T\vecw} - \lambda.
\end{eqnarray}
If we compare the $\vecw$ learning rule (\ref{eq_wruleavg_pca_l2})
with (\ref{eq_wruleavg_pca_sum}) we see that $\vecw^T\matC\vecw$ has
been replaced by $\vecone^T\matC\vecw$ and that the second term has
disappeared. Comparing the $\lambda$ learning rule
(\ref{eq_lruleavg_pca_l2}) with (\ref{eq_lruleavg_pca_sum}) reveals
that these equations differ by a factor $\vecw^T\vecw$: The Rayleigh
quotient is necessary since $\vecw$ is not a (L2) unit vector in the
zero point. However, the Rayleigh quotient is unfortunate since
it requires the computation of $\vecw^T\vecw$. In the vicinity of the
zero point, the Rayleigh quotient and the term
$\vecone^T\matC\vecw$ from the $\vecw$ rule coincide, so we assume
that (\ref{eq_lruleavg_pca_sum}) can be replaced by
\begin{equation}\label{eq_lruleavg_pca_sum_mod}
\dot{\lambda} = \vecone^T\matC\vecw - \lambda.
\end{equation}
This assumption is supported by the fact that similar terms appear in
the coupled SVD rules with constant-sum constraint
(\ref{eq_srule_avg_svd_sum},\ref{eq_rrule_avg_svd_sum}).

The online form of the system
(\ref{eq_wruleavg_pca_sum},\ref{eq_lruleavg_pca_sum_mod}) is
\begin{eqnarray}
\vecwdot &=& \lambda^{-1} \xi (\vecx - (\vecone^T\vecx)\vecw)\\
\dot{\lambda} &=& (\vecone^T\vecx)\xi - \lambda.
\end{eqnarray}

\section{SVD}

\subsection{SVD Objective Function}

To define the objective function for SVD, we introduce the projections
of two different input vectors $\vecx$ (dimension $n$) and $\vecy$
(dimension $m$) onto vectors $\vecv$ and $\vecu$, respectively:
\begin{equation}
\hat{\xi} = \frac{\vecv^T}{\|\vecv\|}\vecx,\quad
\hat{\eta} = \frac{\vecu^T}{\|\vecu\|}\vecy.
\end{equation}
The objective of SVD is to find extrema in $\vecu$ and $\vecv$ of the
covariance of the projection
\begin{eqnarray}
p(\vecu,\vecv) 
&=& E\{\hat{\eta}\hat{\xi}\}\\
&=& E\left\{\frac{\vecu^T}{\|\vecu\|}\vecy
\vecx^T\frac{\vecv}{\|\vecv\|}\right\}\\
&=& \frac{\vecu^T}{\|\vecu\|}E\{\vecy\vecx^T\}\frac{\vecv}{\|\vecv\|}\\
&=& \frac{\vecu^T\matA\vecv}{\|\vecu\|\|\vecv\|} 
= \frac{\vecv^T\matA^T\vecu}{\|\vecu\|\|\vecv\|},
\end{eqnarray}
where $\matA = E\{\vecy\vecx^T\}$ is the cross-covariance matrix of
the distribution formed by vector pairs $(\vecy,\vecx)$.

The derivative of the scalar product of a constant vector with a unit
vector is derived in appendix \ref{app_scalar}; see equation
(\ref{eq_derivative_scalar}). The extrema (stationary points) can be
determined from
\begin{equation}
\ddf{p}{\vecu} =
\frac{\matA\frac{\vecv}{\|\vecv\|}\|\vecu\|
      -\frac{\vecv^T}{\|\vecv\|}\matA^T\vecu\frac{\vecu}{\|\vecu\|}}
{\vecu^T\vecu} = \vecnull
\end{equation}
and
\begin{equation}
\ddf{p}{\vecv} =
\frac{\matA^T\frac{\vecu}{\|\vecu\|}\|\vecv\|
      -\frac{\vecu^T}{\|\vecu\|}\matA\vecv\frac{\vecv}{\|\vecv\|}}
{\vecv^T\vecv} = \vecnull,
\end{equation}
leading to
\begin{eqnarray}
\matA\vecv - \frac{\vecv^T\matA^T\vecu}{\vecu^T\vecu} \vecu &=& \vecnull\\
\matA^T\vecu - \frac{\vecu^T\matA\vecv}{\vecv^T\vecv} \vecv &=& \vecnull.
\end{eqnarray}
We introduce the scalar estimates
\begin{eqnarray}
\sigma &=& \frac{\vecv^T\matA^T\vecu}{\vecu^T\vecu}\label{eq_svd_sigma}\\
\rho   &=& \frac{\vecu^T\matA\vecv}{\vecv^T\vecv}\label{eq_svd_rho}
\end{eqnarray}
and obtain the functions to which the zero finder is applied:
\begin{eqnarray}
\matA\vecv   &=& \sigma \vecu\label{eq_svd_1}\\
\matA^T\vecu &=& \rho \vecv.\label{eq_svd_2}
\end{eqnarray}
The consistency can be checked by inserting (\ref{eq_svd_1}) into
(\ref{eq_svd_sigma}) and (\ref{eq_svd_2}) into
(\ref{eq_svd_rho}). Note that $\sigma$ and $\rho$ only coincide if
$\|\vecu\| = \|\vecv\| = 1$.

\subsection{SVD with Euclidean Weight Vector Norm}

If $\|\vecu\| = \|\vecv\| = 1$, equations (\ref{eq_svd_sigma}) and
(\ref{eq_svd_rho}) coincide and thus we only have a single scalar
estimate $\sigma = \rho$. Moreover, if we guarantee the constraint
$\|\vecu\| = 1$ in the zero point, we automatically ensure that
$\|\vecv\| = 1$: From $\matA\vecv=\sigma\vecu$ we obtain
$\vecu^T\matA\vecv=\sigma\vecu^T\vecu=\sigma$ if $\|\vecu\| = 1$, and
from $\matA^T\vecu=\sigma\vecv$ we obtain
$\vecv^T\matA^T\vecu=\sigma\vecv^T\vecv$; since
$\vecv^T\matA^T\vecu=\vecu^T\matA\vecv$ we can conclude that
$\|\vecv\| = 1$. Therefore we only have to include a single constraint
into our function. This reduction is important as otherwise the
Jacobian would be non-square and could not be inverted.
 
We define the following equation over the vector $\vecz^T = (\vecu^T |
\vecv^T | \sigma)$:
\begin{equation}
\vecf(\vecz) = \vecf(\vecu,\vecv,\sigma) = 
\pmat{
\matA\vecv - \sigma\vecu\\
\matA^T\vecu - \sigma\vecv\\
\half(\vecu^T\vecu - 1)
}.
\end{equation}
%
%
The Jacobian of this system is
\begin{equation}
\matJ(\vecu,\vecv,\sigma) = \ddf{\vecf(\vecz)}{\vecz} = \pmat
{
-\sigma\matI_m & \matA & -\vecu\\
\matA^T & -\sigma\matI_n & -\vecv\\
\vecu^T & \vecnull_n^T & 0
}.
\end{equation}
For the orthogonal transformation we define $\mattU$, the orthogonal
$m\times m$ matrix containing all left singular vectors
$\vectu_i,\;i=1,\ldots,m$, and $\mattV$, the orthogonal $n\times n$
matrix containing all right singular vector $\vectv_i,\;i=1,\ldots,n$,
both sorted such that $|\sigma_1| \gg |\sigma_i|\;\forall i\neq 1$
holds for the corresponding singular values. The transformation matrix
is defined as
\begin{equation}
\matT = \pmat{
\mattU & \matNull_{mn} & \vecnull_m\\
\matNull_{nm} & \mattV & \vecnull_n\\
\vecnull^T_m  & \vecnull^T_n & 1
}.
\end{equation}
We also introduce the $m\times n$ matrix $\matS$ whose first
$\min\{m,n\}$ diagonal elements $\sigma_i$ are the singular values,
sorted as described above. We approximate $\vecu \approx \vectu_1$,
$\vecv \approx \vectv_1$, and $\sigma \approx \sigma_1$. With
$\matA\mattV=\mattU\matS$ and $\matA^T\mattU=\mattV\matS^T$ and the
transformation (\ref{eq_trans_fwd}) we get
\begin{equation}
\matJ^* = \pmat{
-\sigma\matI_m & \matS & -\vece_m\\
\matS^T & -\sigma\matI_n & -\vece_n\\
\vece_m^T & \vecnull_n^T & 0
}.
\end{equation}
We approximate $\matS \approx \sigma \vece_m\vece_n^T$. Using
Gauss-Jordan elimination on the single-element form of $\matJ^*$ we
get
\begin{equation}
{\matJ^*}^{-1} \approx \pmat{
-\sigma^{-1} (\matI_m-\vece_m\vece_m^T) & \matNull_{mn} & \vece_m\\
\half\sigma^{-1}\vece_n\vece_m^T & -\sigma^{-1}(\matI_n-\half\vece_n\vece_n^T)&\vece_n\\
-\half\vece_m^T & -\half\vece_n^T & 0
}.
\end{equation}
The inverse orthogonal transformation (\ref{eq_trans_inv}) leads to
\begin{equation}
\matJ^{-1}(\vecu,\vecv,\sigma) \approx \pmat{
-\sigma^{-1} (\matI_m-\vecu\vecu^T) & \matNull_{mn} & \vecu\\
\half\sigma^{-1}\vecv\vecu^T & -\sigma^{-1}(\matI_n-\half\vecv\vecv^T)&\vecv\\
-\half\vecu^T & -\half\vecv^T & 0
},
\end{equation}
and the Newton zero-finding equation
\begin{equation}
\pmat{\dot{\vecu}\\ \dot{\vecv}\\ \dot{\sigma}} =
-\matJ^{-1}(\vecu,\vecv,\sigma) \vecf(\vecu,\vecv,\sigma)
\end{equation}
gives
\begin{eqnarray}
\dot{\vecu} 
&=& \sigma^{-1} (\matA\vecv - (\vecu^T\matA\vecv)\vecu) 
+ \half(\vecu^T\vecu-1)\vecu\\
\dot{\vecv}
&=& \sigma^{-1} (\matA^T\vecu - (\vecv^T\matA^T\vecu)\vecv)
+ \half(\vecv^T\vecv-1)\vecv\\
\dot{\sigma}
&=& \vecu^T\matA\vecv-\half\sigma(\vecu^T\vecu+\vecv^T\vecv)
\end{eqnarray}
which coincides with the rules derived by \cite{own_Kaiser10}.

The online rules are obtained by replacing $\matA$ by $\vecy\vecx^T$
and introducing the neuron activities $\xi = \vecv^T\vecx$ and $\eta =
\vecu^T \vecy$, which leads to
\begin{eqnarray}
\dot{\vecu}\label{eq_svd_l2_online_u}
&=&
\sigma^{-1} \xi(\vecy - \eta\vecu) 
+ \half(\vecu^T\vecu-1)\vecu\\
\dot{\vecv}\label{eq_svd_l2_online_v}
&=&
\sigma^{-1} \eta(\vecx - \xi\vecv)
+ \half(\vecv^T\vecv-1)\vecv\\
\dot{\sigma}\label{eq_svd_l2_online_sigma}
&=&
\eta\xi - \half\sigma(\vecu^T\vecu+\vecv^T\vecv).
\end{eqnarray}
In the vicinity of the zero point we can further approximate for
$\|\vecu\| \approx 1$ and $\|\vecv\| \approx 1$ such that the second
terms of equation (\ref{eq_svd_l2_online_u}) and
(\ref{eq_svd_l2_online_v}) disappear and equation
(\ref{eq_svd_l2_online_sigma}) turns into
\begin{equation}
\dot{\sigma} = \eta\xi - \sigma.
\end{equation}

\subsection{SVD with Constant Weight Vector Sum}

For the constraint of constant weight vector sums, $\sigma$
(\ref{eq_svd_sigma}) and $\rho$ (\ref{eq_svd_rho}) do not coincide. We
define the following equation over the vector $\vecz^T =
(\vecu^T,\vecv^T,\sigma,\rho)$:
\begin{equation}
\vecf(\vecz) = \vecf(\vecu,\vecv,\sigma,\rho) = \pmat{
\matA\vecv - \sigma\vecu\\
\matA^T\vecu - \rho\vecv\\
\vecone_m^T\vecu - 1\\
\vecone_n^T\vecv - 1
}.
\end{equation}
We obtain a square Jacobian
\begin{equation}
\matJ(\vecu,\vecv,\sigma,\rho) = \ddf{\vecf(\vecz)}{\vecz} = \pmat{
-\sigma\matI_m & \matA & -\vecu & \vecnull_m\\
\matA^T & -\rho\matI_n & \vecnull_n & -\vecv\\
\vecone_m^T & \vecnull_n^T & 0 & 0\\
\vecnull_m^T & \vecone_n^T & 0 & 0
}.
\end{equation}
For the orthogonal transformation we define $\mattU$, the orthogonal
$m\times m$ matrix containing all left singular vectors
$\vectu_i,\;i=1,\ldots,m$, and $\mattV$, the orthogonal $n\times n$
matrix containing all right singular vector $\vectv_i,\;i=1,\ldots,n$,
both sorted according to the corresponding singular values $\mu_i$
obtained for L2 unit-length left and right singular vectors such that
$|\mu_1| \gg |\mu_i|\;\forall i\neq 1$. We introduce the $m\times n$
matrix $\matM$ whose first $\min\{m,n\}$ diagonal elements are the
singular values $\mu_i$ (obtained for L2 unit-length vectors), sorted
as described above. This matrix can be approximated as $\matM \approx
\mu_1 \vece_m \vece_n^T \approx \mu \vece_m \vece_n^T$. We can use the
relationships $\matA\mattV=\mattU\matM$ and
$\matA^T\mattU=\mattV\matM^T$. The transformation matrix is defined as
\begin{equation}\label{eq_trans_svd_sum}
\matT = \pmat{
\mattU & \matNull_{mn} & \vecnull_m & \vecnull_m\\
\matNull_{nm} & \mattV & \vecnull_n & \vecnull_n\\
\vecnull^T_m  & \vecnull^T_n & 1 & 0\\
\vecnull^T_m  & \vecnull^T_n & 0 & 1
}.
\end{equation}
We now have to establish the relationships between the L2 unit-length
vectors in $\mattU$ and $\mattV$ and the weight vectors $\vecu$ and
$\vecv$ which in the zero point are constrained to constant sum:
\begin{eqnarray}
\label{eq_u_ut}
\vecu_i &=& \frac{\vectu_i}{\vecone_m^T\vectu_i},\quad
\vectu_i = \frac{\vecu_i}{\|\vecu_i\|},\quad
\mbox{thus} \quad \|\vecu_i\|\cdot(\vecone_m^T\vectu_i) = 1\\
\label{eq_v_vt}
\vecv_i &=& \frac{\vectv_i}{\vecone_n^T\vectv_i},\quad
\vectv_i = \frac{\vecv_i}{\|\vecv_i\|},\quad
\mbox{thus} \quad \|\vecv_i\|\cdot(\vecone_n^T\vectv_i) = 1
\end{eqnarray}
(and see footnote \ref{fn_w_wt}). We approximate $\vecu \approx
\vecu_1$, $\vecv \approx \vecv_1$, and $\mu \approx \mu_1$. We apply
the transformation (\ref{eq_trans_fwd}), use $\mattU^T\vecu =
\mattU^T\vectu\|\vecu\| \approx \vece_m \|\vecu\|$ and $\mattV^T\vecv
= \mattV^T\vectv\|\vecv\| \approx \vece_n \|\vecv\|$, and get
\begin{equation}
\matJ^* \approx \pmat{
-\sigma\matI_m & \mu\vece_m\vece_n^T & -\vece_m\|\vecu\| & \vecnull_m\\
\mu\vece_n\vece_m^T & -\rho\matI_n & \vecnull_n & -\vece_n\|\vecv\|\\
\vecone_n^T\mattU & \vecnull_n^T & 0 & 0\\
\vecnull_m^T & \vecone_m^T\mattV & 0 & 0
}.
\end{equation}
For the single-element inversion we introduce the vectors
\begin{eqnarray}
\vecs^T &=& \vecone_m^T\mattU = (s_1,\ldots,s_m)\label{eq_s}\\
\vecr^T &=& \vecone_n^T\mattV = (r_1,\ldots,r_n).\label{eq_r}
\end{eqnarray}
The inversion of $\matJ^*$ yields
\newcommand{\veccs}{\check{\vecs}}
\newcommand{\veccr}{\check{\vecr}}
\begin{eqnarray}
&&{\matJ^*}^{-1} \approx \\
&& \nonumber\pmat{
  \sigma^{-1}[s_1^{-1}\vece_m\veccs^T - (\matI_m - \vece_m\vece_m^T)] & 
  \matNull_{mn} & 
  s_1^{-1} \vece_m & 
  \vecnull_m\\
  \matNull_{nm} & 
  \rho^{-1}[r_1^{-1}\vece_n\veccr^T - (\matI_n - \vece_n\vece_n^T)] &
  \vecnull_n &
  r_1^{-1} \vece_n\\
  -\vecs^T &
  \mu \rho^{-1} s_1 r_1^{-1} \veccr^T &
  -\sigma &
  \mu s_1 r_1^{-1}\\
  \mu \sigma^{-1} r_1 s_1^{-1} \veccs^T &
  -\vecr^T &
  \mu r_1 s_1^{-1} &
  -\rho\\
}
\end{eqnarray}
where $\veccs^T=(0,s_2,\ldots,s_m)$ and
$\veccr^T=(0,r_2,\ldots,r_n)$. For the test $\matJ^*{\matJ^*}^{-1} =
\matI_{m+n+2}$, note that $s_1^{-1} = \|\vecu\|$ and $r_1^{-1} =
\|\vecv\|$ which results from equations (\ref{eq_u_ut},\ref{eq_v_vt})
and (\ref{eq_s},\ref{eq_r}).

For the inverse transformation (\ref{eq_trans_inv}) we use the following
relationships:
\begin{eqnarray}
\vecs^T\mattU^T &=& \vecone_m^T\mattU\mattU^T = \vecone_m\\
\vecr^T\mattV^T &=& \vecone_n^T\mattV\mattV^T = \vecone_n\\
\veccs^T\mattU^T &=& (\vecs^T-s_1\vece_m^T) \mattU^T 
                     = \vecone_m^T - \vecone_m^T\vectu\vectu^T\\
\veccr^T\mattV^T &=& (\vecr^T-r_1\vece_n^T) \mattV^T
                     = \vecone_n^T - \vecone_n^T\vectv\vectv^T\\
s_1^{-1} \vectu &=& \vecu\\
r_1^{-1} \vectv &=& \vecv\\
s_1^{-1} &=& \|\vecu\|\\
r_1^{-1} &=& \|\vecv\|
\end{eqnarray}
and obtain
\begin{equation}
\matJ^{-1}(\vecu,\vecv,\sigma,\rho) = \pmat{
\sigma^{-1}(\vecu\vecone_m^T-\matI_m) & \matNull_{mn} & \vecu & \vecnull_m\\
\matNull_{nm} & \rho^{-1}(\vecv\vecone_n^T-\matI_n) & \vecnull_n & \vecv\\
-\vecone_m^T & 
\mu\frac{(\vecv^T\vecv)\vecone_n^T-\vecv^T}{\rho\|\vecu\|\|\vecv\|} &
-\sigma &
\mu\frac{\|\vecv\|}{\|\vecu\|}\\
\mu\frac{(\vecu^T\vecu)\vecone_m^T-\vecu^T}{\sigma\|\vecu\|\|\vecv\|} &
-\vecone_n^T &
\mu\frac{\|\vecu\|}{\|\vecv\|} &
-\rho
}.
\end{equation}
If we apply
\begin{equation}
\pmat{\dot{\vecu}\\\dot{\vecv}\\\dot{\sigma}\\\dot{\rho}} =
-\matJ^{-1}(\vecu,\vecv,\sigma,\rho) \vecf(\vecu,\vecv,\sigma,\rho)
\end{equation}
we obtain
\begin{eqnarray}
\label{eq_urule_avg_svd_sum}
\dot{\vecu} &=& \sigma^{-1}(\matA\vecv-(\vecone_m^T\matA\vecv)\vecu)\\
\label{eq_vrule_avg_svd_sum}
\dot{\vecv} &=& \rho^{-1}(\matA^T\vecu-(\vecone_n^T\matA^T\vecu)\vecv)\\
\label{eq_srule_avg_svd_sum}
\dot{\sigma} &=& \vecone_m^T\matA\vecv - \sigma 
- \frac{\mu}{\rho\|\vecu\|\|\vecv\|}
\left[(\vecv^T\vecv)(\vecone_n^T\matA^T\vecu)-\vecv^T\matA^T\vecu\right]\\
\label{eq_rrule_avg_svd_sum}
\dot{\rho} &=& \vecone_n^T\matA^T\vecu - \rho
- \frac{\mu}{\sigma\|\vecu\|\|\vecv\|}
\left[(\vecu^T\vecu)(\vecone_m^T\matA\vecv)-\vecu^T\matA\vecv\right].
\end{eqnarray}
The last terms of
(\ref{eq_srule_avg_svd_sum},\ref{eq_rrule_avg_svd_sum}) are cumbersome
as they require the computation of L2 lengths of the weight vectors
(which are constrained to unit sum) and need an additional ODE which
estimates $\mu$. From
(\ref{eq_urule_avg_svd_sum},\ref{eq_vrule_avg_svd_sum}) we can
conclude that $\vecone_m^T\matA\vecv = \sigma$ and
$\vecone_n^T\matA^T\vecu = \rho$ are valid in the stationary
point. Since we also have $\vecv^T\matA^T\vecu/(\vecu^T\vecu) =
\sigma$ (\ref{eq_svd_sigma}) and $\vecu^T\matA\vecv/(\vecv^T\vecv) =
\rho$ (\ref{eq_svd_rho}), and $\vecu^T\matA\vecv=\vecv^T\matA^T\vecu$,
we see that the terms are at least small in the vicinity of the
stationary point; however they are not necessarily smaller than the
remaining terms. It is therefore not obvious how the approximations
\begin{eqnarray}
\label{eq_srule_avg_svd_sum_mod}
\dot{\sigma} &=& \vecone_m^T\matA\vecv - \sigma\\
\label{eq_rrule_avg_svd_sum_mod}
\dot{\rho} &=& \vecone_n^T\matA^T\vecu - \rho,
\end{eqnarray}
where the last terms are omitted, affect the behavior of
(\ref{eq_srule_avg_svd_sum},\ref{eq_rrule_avg_svd_sum}). However, at
least the system
(\ref{eq_urule_avg_svd_sum},\ref{eq_vrule_avg_svd_sum},\ref{eq_srule_avg_svd_sum_mod},\ref{eq_rrule_avg_svd_sum_mod})
has the proper stationary points $\matA\vecv=\sigma\vecu$,
$\matA^T\vecu=\rho\vecv$, $\vecone_m^T\vecu=\vecone_n^T\vecv=1$. The
stability analysis of this system is presented in appendix
\ref{app_stab_svd_sum}.

The online rules are obtained by replacing $\matA$ by $\vecy\vecx^T$
and introducing the neuron activities $\xi = \vecv^T\vecx$ and $\eta =
\vecu^T \vecy$, which leads to
\begin{eqnarray}
\dot{\vecu} &=& \sigma^{-1}\xi(\vecy-(\vecone_m^T\vecy)\vecu)\\
\dot{\vecv} &=& \rho^{-1}\eta(\vecx-(\vecone_n^T\vecx)\vecv)\\
\dot{\sigma} &=& (\vecone_m^T\vecy)\xi - \sigma\\
\dot{\rho} &=& (\vecone_n^T\vecx)\eta -\rho.
\end{eqnarray}

\section{Discussion}

\subsection{Newton Zero-Finding Framework}\label{sec_disc_newton_zero}

Deriving coupled learning rules from either the Newton optimization
framework or the Newton zero-finding framework leads to rules which
are similar to those derived by Taylor expansions of normalization for
small learning rates \cite[as done by][]{nn_Oja82}. At least in
simplified form and for the {\em principal} component case, the
coupling always takes the form of multiplying the ODE of the vector
estimate by an inverse scalar estimate (eigenvalue, singular
value). This may raise the question whether the Newton approach is too
complicated compared to the Taylor approach. There are two arguments
in favor of the Newton approach:
\begin{itemize}
\item The Taylor approach only produces learning rules for {\em
  principal} component estimates. As shown by \cite{own_Moeller04a},
  the Newton framework can also be used to derive {\em minor}
  component rules by approximating the Hessian or Jacobian in the
  vicinity of this stationary or zero point (but no online rules can
  directly be derived for this case as the {\em inverse} covariance
  matrix appears in the solution). We can conclude that the Newton
  approach is more general.
\item Additional terms appear in the update equations derived from the
  Newton approach, such as the last term in equation
  (\ref{eq_wruleavg_pca_l2}). The terms are required to have
  approximately unit convergence speed from all directions. Leaving
  them out leads to a different convergence speed in one direction
  \cite[]{own_Moeller04a} (however, no effect of this difference is
  apparent in simulations). Therefore the rule-of-thumb ``derive from
  Taylor approach and multiply be inverse scalar estimate'' is only an
  approximation.
\end{itemize}
Nevertheless, it is somewhat worrying that after rather complex
derivations (approximation of the Jacobian / Hessian, orthogonal
transformation, inversion of Hessian / Jacobian, inverse orthogonal
transformation, simplification of resulting ODEs) we obtain quite simple
learning rules. This may indicate that there is a simpler way to
derive these rules or some generalization for the given class of
problems (PCA, SVD, GPCA).

The advantages of the Newton {\em zero-finding} framework over the the
Newton {\em optimization} framework could be demonstrated in this
paper: a clear derivation starting from an objective function related
to the problem at hand (rather than from a ``designed'' information
criterion with no explanatory value) and the possibility to add
arbitrary constraints on the vector estimates (rather than just
Euclidean constraints implicitly embedded in the information
criterion). The constant-sum constraint was deliberately chosen in
this work as it allows to obtain neurons which specialize to represent
the conjunction (logical ``and'') of binary (0/1) inputs. Note that for
Euclidean constraints, the zero-finding and the optimization framework
produce the same learning rules.

One important step in the Newton zero-finding framework is the
orthogonal transformation of the Jacobian (which allows an approximation in
the vicinity of the desired solution). The orthogonal transformation
requires orthogonal matrices with estimates of the eigenvectors /
singular vectors, thus these vectors have Euclidean unit length. In
contrast, different constraints are imposed on the vector estimates in
the ODEs. Transformations need to be introduced to interrelate between
both types of vectors (equations (\ref{eq_w_wt}), (\ref{eq_u_ut}),
(\ref{eq_v_vt})). In the SVD constant-sum case, this unfortunately
introduces the singular value estimate $\mu$ into the equations which
relates to the Euclidean unit-length vectors. This variable survives
into the update equations of the two other singular value estimates
$\sigma$ and $\rho$ (equations
(\ref{eq_srule_avg_svd_sum},\ref{eq_rrule_avg_svd_sum})). So far we
have no suggestion how this can be avoided.

In some cases, the Newton zero-finding framework leads to solutions
which are awkward in an implementation. In the PCA constant-sum case,
the update equation for the eigenvalue (\ref{eq_lruleavg_pca_sum})
includes the squared Euclidean norm of the eigenvector estimate
($\vecw^T\matC\vecw/\vecw^T\vecw$). It is more convenient to replace
this by $\vecone^T\matC\vecw$ (equation
(\ref{eq_lruleavg_pca_sum_mod})) as this term also appears in the
update equation for the vector estimate
(\ref{eq_wruleavg_pca_sum}). Surprisingly, the desired terms appear in
the update equations of the singular value estimates in the SVD
constant-sum case (first terms of equations
(\ref{eq_srule_avg_svd_sum},\ref{eq_rrule_avg_svd_sum})).

\subsection{Limitation and Alternative Lagrange-Newton Framework}
\label{sec_disc_lagrange_newton}

A note of caution has to be added here. The standard approach to solve
an optimization problem under a given constraint would be to use the
method of Lagrange multipliers: An optimization criterion is combined
with all constraint equations multiplied by a vector of Lagrange
multipliers. Here, in contrast, we do not consider the optimization
criterion but its {\em unconstrained} optimum given by its derivative
(PCA: (\ref{eq_pca}), SVD: (\ref{eq_svd_1},\ref{eq_svd_2})). The
system of equations obtained by combining the derivative of the
optimization criterion with the constraint equations only leads to a
solution, {\em if the constraints intersect the unconstrained
  optimum}. In all four cases described here, the unconstrained
optimum allows for arbitrary vectors lengths, so the vector-length
constraints always intersect the unconstrained optimum. In other cases
where this condition is not fulfilled, the suggested zero-finding
framework will fail to provide a solution. This is a clear limitation
of the Newton zero-finding framework.

Actually it should be possible to derive the same learning rules from
a Lagrange-Newton framework. In the Lagrange-Newton framework, the
Lagrange-multiplier variables are considered in the Newton
step.\footnote{A Newton step is actually necessary, since the
  solutions of the Lagrange equations are typically saddle points,
  thus a gradient descent or ascent would not be sufficient. Applying
  a Newton descent turns the saddle into an attractor. See appendix
  \ref{app_saddle} for an example. I couldn't find a proof so far.} In
the following we sketch the solution for the first case, PCA with
Euclidean constraint. The Lagrange-multiplier equation is
\begin{equation}
J(\vecw, \alpha) = \half \vecw^T \matC \vecw - \half \alpha (\vecw^T\vecw - 1),
\end{equation}
where $\alpha$ is the Lagrange multiplier. The derivatives are
\begin{eqnarray}
  \ddf{J}{\vecx}  &=& \matC \vecw - \alpha \vecw\\
  \ddf{J}{\alpha} &=& -\half(\vecw^T\vecw - 1).
\end{eqnarray}
We see that, except for the sign of the second equation, this
coincides with (\ref{eq_f_pca_l2}). We obtain the Hessian
\begin{equation}
  \matH(\vecw,\alpha) =
  \pmat{ \matC - \alpha \matI & -\vecw\\ -\vecw^T & 0 },
\end{equation}
and, in a similar way as in section \ref{sec_pca_l2}, the approximated
inverse
\begin{equation}
  \matH^{-1}(\vecw,\alpha) \approx
  \pmat{ \alpha^{-1}(\vecw\vecw^T-\matI) & -\vecw\\ -\vecw^T & 0 }.
\end{equation}
A Newton descent\footnote{Note that regardless of whether the
  criterion should be maximized or minimized, it is always a Newton
  {\em descent} step. This is different from following a gradient:
  Maximizing a criterion needs a gradient {\em ascent}, minimizing a
  gradient {\em descent}.} leads to the same learning rule ODEs as
(\ref{eq_wruleavg_pca_l2}, \ref{eq_lruleavg_pca_l2}), except for using
the name $\alpha$ instead of $\lambda$.

The derivation of the other three cases should be similar, but hasn't
been performed yet.

\section{Conclusion}

Despite some open problems mentioned in the discussion, the value of
the novel Newton zero-finding framework as a way to systematically
derive coupled learning rules with arbitrary vector constraints from
objective functions has been demonstrated. The four examples
elaborated in this paper can serve as a guideline for the derivation
of learning rules for other problems (such as GPCA).

\section{Acknowledgements}

Thanks to Alexander Kaiser for corrections of the manuscript.

\section{Changes}
\label{sec_changes}

March 13, 2017: Updated reference \cite[]{nn_Feng17}.

March 14, 2017: Corrected reference \cite[]{nn_Feng17}.

April 15, 2019: Added subsection \ref{sec_disc_lagrange_newton} to
discussion (original discussion now in subsection
\ref{sec_disc_newton_zero}). Added appendix \ref{app_saddle} with
example of saddle point in Lagrange-multiplier equation.

March 25, 2020: arXiv version: different title page, moved appendix

\trarxiv{%
  \bibliographystyle{/home/moeller/bst/plainnatsfnnm}
  \bibliography{/home/moeller/bib/nn17,/home/moeller/bib/own14}

\begin{thebibliography}{9}
\expandafter\ifx\csname natexlab\endcsname\relax\def\natexlab#1{#1}\fi

\bibitem[Chen and Chang(1995)]{nn_Chen95}
L.-H. Chen and S.~Chang.
\newblock An adaptive learning algorithm for principal component analysis.
\newblock {\em IEEE Transactions on Neural Networks}, 6\penalty0 (5):\penalty0
  1255--1263, 1995.

\bibitem[Diamantaras and Kung(1994)]{nn_Diamantaras94}
K.~I. Diamantaras and S.-Y. Kung.
\newblock Cross-correlation neural network models.
\newblock {\em IEEE Transactions on Signal Processing}, 42\penalty0
  (11):\penalty0 3218--3223, 1994.

\bibitem[Feng et~al.(2016)Feng, Kong, Duan, and Ma]{nn_Feng16a}
X.~Feng, X.~Kong, Z.~Duan, and H.~Ma.
\newblock Adaptive generalized eigen-pairs extraction algorithms and their
  convergence analysis.
\newblock {\em IEEE Transactions on Signal Processing}, 64\penalty0
  (11):\penalty0 2976--2989, 2016.

\bibitem[Feng et~al.(2017)Feng, Kong, Ma, and Liu]{nn_Feng17}
X.~Feng, X.~Kong, H.~Ma, and H.~Liu.
\newblock Unified and coupled self-stabilizing algorithms for minor and
  principal eigen-pairs extraction.
\newblock {\em Neural Processing Letters}, 45\penalty0 (1):\penalty0 197--222,
  2017.
\newblock doi:10.1007/s11063-016-9520-3.

\bibitem[Hou and Chen(2006)]{nn_Hou06}
L.~Hou and T.~Chen.
\newblock Online algorithm of coupled principal (minor) component analysis.
\newblock {\em Journal of Fudan University (Natural Science)}, 45\penalty0
  (2):\penalty0 158--169, 2006.

\bibitem[Kaiser et~al.(2010)Kaiser, Schenck, and M\"oller]{own_Kaiser10}
A.~Kaiser, W.~Schenck, and R.~M\"oller.
\newblock Coupled singular value decomposition of a cross covariance matrix.
\newblock {\em International Journal of Neural Systems}, 20\penalty0
  (4):\penalty0 293--318, 2010.

\bibitem[M\"oller and K\"onies(2004)]{own_Moeller04a}
R.~M\"oller and A.~K\"onies.
\newblock Coupled principal component analysis.
\newblock {\em IEEE Transactions on Neural Networks}, 15\penalty0 (1):\penalty0
  214--222, 2004.

\bibitem[Nguyen and Yamada(2013)]{nn_Nguyen13}
T.~D. Nguyen and I.~Yamada.
\newblock Adaptive normalized quasi-newton algorithms for extraction of
  generalized eigen-pairs and their convergence analysis.
\newblock {\em IEEE Transactions on Signal Processing}, 61\penalty0
  (6):\penalty0 1404--1418, 2013.

\bibitem[Oja(1982)]{nn_Oja82}
E.~Oja.
\newblock A simplified neuron model as principal component analyzer.
\newblock {\em Journal of Mathematical Biology}, 15:\penalty0 267--273, 1982.

\end{thebibliography}
}{%

}


\appendix

\section{Derivative of the Rayleigh Quotient}\label{app_rayleigh}

The vector derivative of the Rayleigh quotient
\begin{equation}
\df{\vecx} \frac{\vecx^T \matA \vecx}{\vecx^T \vecx}
\end{equation}
is obtained by computing the scalar derivative
\begin{equation}
\df{x_j} \frac{\vecx^T \matA \vecx}{\vecx^T \vecx} =
\df{x_j} \frac{\sum_{l,m} x_l A_{lm} x_m}{\sum_k x_k^2}.
\end{equation}
The derivative of the numerator $u$ is obtained from the product rule
\begin{eqnarray}
u' 
&=& \df{x_j} \sum_{l,m} x_l A_{lm} x_m\\
&=& \sum_{l,m} \delta_{lj} A_{lm} x_m + \sum_{l,m} x_l A_{lm} \delta_{mj}
\end{eqnarray}
where $\delta$ is Kronecker's delta and $\partial x_i / \partial x_j =
\delta_{ij}$ is used. If a sum runs over one index of $\delta$, the
sum disappears and its index is replaced everywhere by the other index
of $\delta$, which here leads to
\begin{equation}
u'
= \sum_m A_{jm} x_m + \sum_l x_l A_{lj}
= (\matA\vecx)_j + (\matA^T\vecx)_j.
\end{equation}
The derivative of the denominator $v$ is
\begin{equation}
v' = 2 x_j = 2 (\vecx)_j.
\end{equation}
The derivative of $u/v$ given by $(u'v - v'u) / v^2$ is
\begin{equation}
\df{x_j}\frac{u}{v} =
\frac{\left[(\matA\vecx)_j + (\matA^T\vecx)_j\right] 
(\vecx^T\vecx) - 2 (\vecx)_j (\vecx^T\matA\vecx)}
{(\vecx^T\vecx)^2}
\end{equation}
which, in vector form, is
\begin{eqnarray}
\df{\vecx} \frac{\vecx^T \matA \vecx}{\vecx^T\vecx}
&=&
\frac{1}{\vecx^T\vecx}
\left[
(\matA + \matA^T)\vecx - 2\vecx\frac{\vecx^T\matA\vecx}{\vecx^T\vecx}
\right].
\end{eqnarray}
For symmetric $\matA$, i.e. $\matA = \matA^T$, we obtain the special form
\begin{equation}\label{eq_derivative_rayleigh_symm}
\df{\vecx} \frac{\vecx^T \matA \vecx}{\vecx^T\vecx}
= 
\frac{2}{\vecx^T\vecx}
\left[
\matA\vecx - \vecx\frac{\vecx^T\matA\vecx}{\vecx^T\vecx}
\right].
\end{equation}

\section{Derivative of a Scalar Product with a Unit Vector}\label{app_scalar}

The vector derivative of
\begin{equation}
\df{\vecx}\veca^T\frac{\vecx}{\|\vecx\|}
\end{equation}
is obtained by computing the scalar derivative
\begin{equation}
\df{x_i}\veca^T\frac{\vecx}{\|\vecx\|} =
\df{x_i}\frac{\sum_j a_j x_j}{\sqrt{\sum_k x_k^2}}.
\end{equation}
The derivative of the numerator $u$ is
\begin{equation}
u' = \df{x_i} \sum_j a_j x_j = \sum_j a_j \delta_{ij} = a_i,
\end{equation}
the derivative of the denominator $v$ is
\begin{eqnarray}
v'
&=& \df{x_i} \left(\sum_k x_k^2\right)^{\half}
 =  \half (\vecx^T\vecx)^{-\half} \df{x_i} \sum_k x_k^2
 =  \half \|\vecx\|^{-1} \sum_k \df{x_j} x_k^2\\
&=& \half \|\vecx\|^{-1} \sum_k 2 x_k \delta_{ik}
 =  \|\vecx\|^{-1} x_i,
\end{eqnarray}
such that the derivative of $u/v$ given by $(u'v - v'u) / v^2$ is
\begin{equation}
\df{x_i} \frac{u}{v} =
\frac{a_i \|\vecx\| - (\veca^T\vecx) \|\vecx\|^{-1} x_i}
{\vecx^T\vecx}.
\end{equation}
In vector form we obtain
\begin{equation}\label{eq_derivative_scalar}
\df{\vecx}\veca^T\frac{\vecx}{\|\vecx\|} = 
\frac{\veca \|\vecx\| - (\veca^T\vecx) \frac{\vecx}{\|\vecx\|}}
{\vecx^T\vecx}.
\end{equation}

\section{Stability Analysis of SVD with Constant Weight Vector Sum}
\label{app_stab_svd_sum}

We analyze the stability of the ODE system
\begin{eqnarray}
\dot{\vecu} &=& \sigma^{-1}(\matA\vecv-(\vecone_m^T\matA\vecv)\vecu)\\
\dot{\vecv} &=& \rho^{-1}(\matA^T\vecu-(\vecone_n^T\matA^T\vecu)\vecv)\\
\dot{\sigma} &=& \vecone_m^T\matA\vecv - \sigma\\
\dot{\rho} &=& \vecone_n^T\matA^T\vecu - \rho.
\end{eqnarray}
The stationary points of this system are characterized by the equations
\begin{eqnarray}
\matA\vecv &=& \sigma\vecu\\
\matA^T\vecv &=& \rho\vecv\\
\sigma &=& \vecone_m^T\matA\vecv\\
\rho &=& \vecone_n^T\matA^T\vecu\\
\vecone_m^T\vecu &=& 1\\
\vecone_n^T\vecv &=& 1.
\end{eqnarray}
The Jacobian of the ODE system is
\begin{scriptsize}
\begin{equation}
\matJ =
\pmat{
-\sigma^{-1}(\vecone_m^T\matA\vecv)\matI_m &
\sigma^{-1}(\matA-\vecu\vecone_m^T\matA) &
-\sigma^{-2}(\matA\vecv-(\vecone_m^T\matA\vecv)\vecu) &
\vecnull_m\\
\rho^{-1}(\matA^T-\vecv\vecone_n^T\matA^T) &
-\rho^{-1}(\vecone_n^T\matA^T\vecu)\matI_n &
\vecnull_n &
-\rho^{-2}(\matA^T\vecu-(\vecone_n^T\matA^T\vecu)\vecv)\\
\vecnull_m^T &
\vecone_m^T\matA &
-1 &
0\\
\vecone_n^T\matA^T &
\vecnull_n^T &
0 &
-1
}.
\end{equation}
\end{scriptsize}%
At the stationary points, the Jacobian turns into 
\begin{equation}
\matJ = \pmat{
-\matI_m &
\sigma^{-1}(\matA-\vecu\vecone_m^T\matA) &
\vecnull_m &
\vecnull_m\\
\rho^{-1}(\matA^T-\vecv\vecone_n^T\matA^T) &
-\matI_n &
\vecnull_n &
\vecnull_n\\
\vecnull_m^T &
\vecone_m^T\matA &
-1 &
0\\
\vecone_n^T\matA^T &
\vecnull_n^T &
0 &
-1
}.
\end{equation}
We analyze the eigenvalues of the Jacobian at the stationary
points. Eigenvalues are invariant under similarity transformations
(and thus also under orthogonal transformations). We apply the
orthogonal transformation (\ref{eq_trans_fwd}) with the transformation
matrix (\ref{eq_trans_svd_sum}). Using the relationships
$\matA\mattV=\mattU\matM$ and $\matA^T\mattU=\mattV\matM^T$ (where
$\matM$ contains the $\min\{m,n\}$ singular values $\mu_i$ with
respect to L2 unit length vectors on its main diagonal), introducing
$\vecs$ from (\ref{eq_s}) and $\vecr$ from (\ref{eq_r}), and
considering $\mattU^T\vecu = \vece_m \|\vecu\|$ and $\mattV^T\vecv =
\vece_n \|\vecv\|$, we obtain
\begin{equation}
\matJ = \pmat{
-\matI_m &
\sigma^{-1}(\matM - \|\vecu\|\vece_m\vecs^T\matM) &
\vecnull_m &
\vecnull_m\\
\rho^{-1}(\matM^T - \|\vecv\|\vece_n\vecr^T\matM^T) &
-\matI_n &
\vecnull_n &
\vecnull_n\\
\vecnull_m^T &
\vecs^T\matM &
-1 &
0\\
\vecr^T\matM^T &
\vecnull_n^T &
0 &
-1
}.
\end{equation}
Eigenvalues of the transformed Jacobian are obtained from the
characteristic equation
\begin{equation}
\det\{\matJ - \lambda \matI_{m+n+2}\} = 0.
\end{equation}
We need to analyze the determinant
\begin{small}
\begin{equation}
\left|\pmat{
-\matI_m (\lambda+1)&
\sigma^{-1}(\matM - \|\vecu\|\vece_m\vecs^T\matM) &
\vecnull_m &
\vecnull_m\\
\rho^{-1}(\matM^T - \|\vecv\|\vece_n\vecr^T\matM^T) &
-\matI_n (\lambda+1)&
\vecnull_n &
\vecnull_n\\
\vecnull_m^T &
\vecs^T\matM &
-(\lambda+1) &
0\\
\vecr^T\matM^T &
\vecnull_n^T &
0 &
-(\lambda+1)
}\right|.
\end{equation}
\end{small}%
We see that the upper right block of size $(m+n)\times 2$ is a zero
matrix, therefore the determinant reduces to the product of the
determinants of the blocks on the main diagonal:
\begin{equation}
= (-\lambda-1)^2
\left|\pmat{
-\matI_m (\lambda+1)&
\sigma^{-1}(\matM - \|\vecu\|\vece_m\vecs^T\matM)\\
\rho^{-1}(\matM^T - \|\vecv\|\vece_n\vecr^T\matM^T) &
-\matI_n (\lambda+1)
}\right|.
\end{equation}
To the four blocks we now apply one of the following equations:
\begin{eqnarray}
\det\pmat{\matA & \matB\\\matC & \matD}
\label{eq_det1}
&=& \det\matA \cdot \det\{\matD - \matC\matA^{-1}\matB\}\\
\label{eq_det2}
&=& \det\matD \cdot \det\{\matA - \matB\matD^{-1}\matC\}.
\end{eqnarray}
We assume $\min\{m,n\} = n$ and apply (\ref{eq_det1}) since this
guarantees that the term $\matM^T\matM$ appearing in the equations
below is a full diagonal matrix (if $\mu_i\neq 0\,\forall
i=1,\ldots,n$); if $\min\{m,n\}=m$, we could apply (\ref{eq_det2}) and
have the guarantee that $\matM\matM^T$ is a full diagonal matrix. In
our case we see that
\begin{eqnarray}
\det\matA &=& (-\lambda-1)^m\\
\matA^{-1} &=& -(\lambda+1)^{-1}\matI_m\\
\matC\matA^{-1}\matB &=& -(\lambda+1)^{-1}\matC\matB.
\end{eqnarray}
We determine
\begin{scriptsize}
\begin{eqnarray}
\matC\matB
&=& 
\rho^{-1}\sigma^{-1}
(\matM^T - \|\vecv\|\vece_n\vecr^T\matM^T)
(\matM - \|\vecu\|\vece_m\vecs^T\matM)\\
&=&
\rho^{-1}\sigma^{-1}
(\matM^T\matM 
- \|\vecv\|\vece_n\vecr^T\matM^T\matM
- \|\vecu\|\matM^T\vece_m\vecs^T\matM 
+ \|\vecv\|\|\vecu\|\vece_n\vecr^T\matM^T\vece_m\vecs^T\matM).
\end{eqnarray}
\end{scriptsize}%
We now apply $\matM^T\vece_m = \mu_1\vece_n$ and $\vecr^T\vece_n =
r_1$ and obtain
\begin{eqnarray}
\matC\matB 
&=&
\rho^{-1}\sigma^{-1}
\left[\left(\matI_n - \|\vecv\|\vece_n\vecr^T\right)\matM^T\matM
- \mu_1\|\vecu\|\left(1-r_1\|\vecv\|\right)\vece_n\vecs^T\matM\right].
\end{eqnarray}
In the matrices $\vece_n\vecr^T(\matM^T\matM)$ and
$\vece_n\vecs^T\matM$, only the top row is occupied by non-zero
elements. Except for the top-row element on the main diagonal, these
elements are irrelevant to the determinant $\det\{\matD +
(\lambda+1)^{-1}\matC\matB\}$ (which is seen immediately if the
determinant is developed along the first column). The top-left element
of $\matC\matB$ is
$\rho^{-1}\sigma^{-1}\mu_1^2(1-s_1\|\vecu\|)(1-r_1\|\vecv\|)$, the
remaining main diagonal is occupied by $\rho^{-1}\sigma^{-1}\mu_i^2$
for $i = 2\ldots n$.

We also have from (\ref{eq_svd_sigma},\ref{eq_svd_rho})
\begin{eqnarray}
\sigma 
&=& \frac{\vecv^T\matA^T\vecu}{\vecu^T\vecu}
= \frac{\vecu^T\matA\vecv}{\vecu^T\vecu}\\
\rho
&=& \frac{\vecu^T\matA\vecv}{\vecv^T\vecv},
\end{eqnarray}
and the singular values with respect to the L2 unit-length vectors are
\begin{equation}
\mu = \frac{\vecu^T}{\|\vecu\|}\matA\frac{\vecv}{\|\vecv\|},
\end{equation}
from which we conclude that $\sigma\rho=\mu^2$.

From this we get
\begin{eqnarray}
\det\{\matJ - \lambda\matI_{m+n+2}\} 
&=&\nonumber
(-\lambda-1)^{m+2}\\
&\cdot&\nonumber
\left[-(\lambda+1)+(\lambda+1)^{-1}\frac{\mu_1^2}{\mu^2}
(1-s_1\|\vecu\|)(1-r_1\|\vecv\|)\right]\\
&\cdot&
\prod_{j=2}^{n}\left[-(\lambda+1)+(\lambda+1)^{-1}\frac{\mu_j^2}{\mu^2}\right]
\end{eqnarray}
and thus the eigenvalues (arranged in the same order as the factors above)
\begin{eqnarray}
\lambda 
&=& 
-1\\
\label{eq_eigen2}
\lambda 
&=& 
-1 \pm \frac{|\mu_1|}{|\mu|} \sqrt{(1-s_1\|\vecu\|)(1-r_1\|\vecv\|)}\\
\label{eq_eigen3}
\lambda
&=&
-1 \pm \frac{|\mu_j|}{|\mu|}\quad j = 2,\ldots n.
\end{eqnarray}
In the following we analyze the stability of the different stationary
points. For that we assume that $|\mu_1|\gg|\mu_2|>\ldots>|\mu_n|>0$.

\paragraph{\textrm{Principal singular triple ($i = 1$)}}

For the principal singular triple
$\vecu=\vecu_1,\vecv=\vecv_1,\mu=\mu_1$ we have $s_1\|\vecu_1\| = 1$
and $r_1 \|\vecv_1\| = 1$ and $|\mu_1| \gg |\mu_j|$ for $j = 2,\ldots,n$,
so we get the eigenvalues
\begin{eqnarray}
\lambda 
&=& 
-1\\
\lambda 
&=& 
-1\\
\lambda
&=&
-1 \pm \frac{|\mu_j|}{|\mu_1|}\approx -1\quad j = 2,\ldots n.
\end{eqnarray}
We see that this stationary point is an attractor (the system is
stable) and that the convergence speed in all eigendirections is
approximately the same ($-1$).

\paragraph{\textrm{Minor singular triples ($i = 3,\ldots,n$)}}

For singular triples $\vecu=\vecu_i,\vecv=\vecv_i,\mu=\mu_i$ for $i =
3,\ldots,n$ we always have an index $j \in \{2,\ldots,n\}$ where
$|\mu_j| > |\mu_i|$ such that equation (\ref{eq_eigen3})
\begin{equation}
\lambda = -1 \pm \underbrace{\frac{|\mu_j|}{|\mu_i|}}_{>1},
\end{equation}
results in one positive eigenvalue, making the stationary point
instable (saddle point).

\paragraph{\textrm{Second singular triple ($i = 2$)}}

For the singular triple $\vecu=\vecu_2,\vecv=\vecv_2,\mu=\mu_2$,
equation (\ref{eq_eigen3}) gives negative eigenvalues for $j =
3,\ldots,n$ and $\lambda = -2, \lambda = 0$ for $j = 2$. So we have
one semistable eigenvalue (0). 

We analyze whether equation (\ref{eq_eigen2}) gives an unstable
eigenvalue (so we can leave the semistable eigenvalue aside):
\begin{eqnarray}
\lambda 
&=&
-1 \pm \frac{|\mu_1|}{|\mu_2|} 
\sqrt{(1-s_1\|\vecu_2\|)(1-r_1\|\vecv_2\|)}\\
&=&
-1 \pm \frac{|\mu_1|}{|\mu_2|} 
\sqrt{\left(1-\frac{\|\vecu_2\|}{\|\vecu_1\|}\right)
\left(1-\frac{\|\vecv_2\|}{\|\vecv_1\|}\right)}.
\end{eqnarray}
The factors under the square root are independent of each other
(depending on the data) and thus can have the same or different
signs. If they have different signs, the eigenvalue is complex and has
a negative real value (stable). If they have the same sign, the
eigenvalue is real, but its sign is not obvious: Even though the first
factor is large ($|\mu_1|/|\mu_2|\gg 1$), the second factor obtained
from the square root can be small. Thus equation (\ref{eq_eigen2})
allows no data-independent stability judgment.

This leaves us with the semistable eigenvalue of $0$ from equation
(\ref{eq_eigen3}). This is called a ``non-hyperbolic fixed
point''.\footnote{Scholarpedia entry ``Equilibrium'',
  \url{www.scholarpedia.org/article/Equilibrium}.} We need to analyze
the non-linear terms of the ODE system in the vicinity of this fixed
point.

\section{Example of a Saddle Point in a Lagrange-Multiplier Equation}
\label{app_saddle}

Consider the following Lagrange-multiplier criterion for $\vecx = (x,
y)$,
\begin{equation}
J(\vecx, \alpha) = \half \vecx^T\vecx + \alpha (\vecone^T \vecx - 1).
\end{equation}
The derivatives are
\begin{eqnarray}
  \ddf{J}{\vecx}  &=& \vecx + \alpha \vecone\\
  \ddf{J}{\alpha} &=& \vecone^T \vecx - 1.
\end{eqnarray}
The Hessian of second derivatives is
\begin{equation}
  \matH
  = \pmat{\matI & \vecone\\ \vecone^T & 0}
  = \pmat{1 & 0 &1\\ 0 & 1 & 1 \\ 1 & 1 & 0}.
\end{equation}
Octave gives the eigenvalues of $\matH$ as $(-1, 1, 2)$, thus the
solution appears at a saddle point.

\end{document}